\documentclass{svproc}
\usepackage{cite}
\usepackage{nameref,hyperref}
\usepackage{changepage}
\usepackage{array}
\usepackage{url}
\usepackage{graphicx} 
\usepackage{amsmath}   
\usepackage{tabularx}

\begin{document}
\mainmatter              % start of a contribution
\title{Data-Driven Machine Learning Approaches for Predicting
In-Hospital Sepsis Mortality}
\titlerunning{Predicting Sepsis Mortality with Data-Driven ML}  % abbreviated title 

\author{Arseniy Shumilov\inst{1} \and Yueting Zhu\inst{1} \and Negin Ashrafi\inst{1} \and Armin Abdollahi\inst{1} \and Greg Placencia\inst{2} \and Kamiar Alaei\inst{3} \and Maryam Pishgar\inst{1}}
\authorrunning{Shumilov et al.} % abbreviated author list (for running head)
% \tocauthor{Arseniy Shumilov, Yueting Zhu, Negin Ashrafi, Armin Abdollahi, Greg Placencia, Kamiar Alaei, Maryam Pishgar}
%
\institute{
University of Southern California, Los Angeles, USA\inst{1} \and
California State Polytechnic University, Pomona, USA\inst{2} \and
California State University, Long Beach, USA\inst{3} \newline
\email{pishgar@usc.edu}
}

\maketitle
\begin{abstract} 
Sepsis is a severe condition responsible for many deaths in the United States and worldwide, making accurate prediction of outcomes crucial for timely and effective treatment. Previous studies employing machine learning faced limitations in feature selection and model interpretability, reducing their clinical applicability. This research aimed to develop an interpretable and accurate machine learning model to predict in-hospital sepsis mortality, addressing these gaps. Using ICU patient records from the MIMIC-III database, we extracted relevant data through a combination of literature review, clinical input refinement, and Random Forest-based feature selection, identifying the top 35 features. Data preprocessing included cleaning, imputation, standardization, and applying the Synthetic Minority Over-sampling Technique (SMOTE) to address class imbalance, resulting in a dataset of 4,683 patients with 17,429 admissions. Five models—Random Forest, Gradient Boosting, Logistic Regression, Support Vector Machine, and K-Nearest Neighbor—were developed and evaluated. The Random Forest model demonstrated the best performance, achieving an accuracy of 0.90, AUROC of 0.97, precision of 0.93, recall of 0.91, and F1-score of 0.92. These findings underscore the potential of data-driven machine learning approaches to improve critical care, offering clinicians a powerful tool for predicting in-hospital sepsis mortality and enhancing patient outcomes.
\end{abstract}

\section{Introduction}

Sepsis is a severe, life-threatening condition characterized by organ dysfunction due to a dysregulated host response to infection ~\cite{bib1}. Prompt identification and effective management can significantly reduce adverse outcomes. In-hospital sepsis mortality is a major issue in critical care due to its high morbidity and mortality rates ~\cite{bib2,bib3,bib4}. Research shows that in-hospital mortality can reach up to 19.27\% with pulmonary sepsis ~\cite{bib5}. A global sepsis report from 2017 estimated 48.9 million cases, with a 95\% confidence interval of 38.9 to 62.9 million cases, resulting in 11 million deaths, representing a 19.7\% fatality rate ~\cite{bib6}. \newline

Despite advances in medical care, early diagnosis and treatment of sepsis remain challenging due to the condition's complexity and variability in patient presentation \cite{bib7}. The use of machine learning models in predicting sepsis outcomes has shown promise in recent years, enabling the identification of at-risk patients and potentially improving clinical interventions \cite{bib8,bib9,bib10}. However, many existing models lack comprehensibility and utilize suboptimal feature selection methods, which can limit their practical applicability in clinical settings. Addressing these limitations by developing interpretable and robust predictive models is crucial to improving the management of sepsis and accurately predicting associated mortality rates. This study aims to fill this gap by leveraging advanced machine learning techniques to create a model that not only achieves high predictive accuracy but is also easily interpretable by healthcare professionals, thus facilitating timely and effective clinical decision-making. \newline

This study harnesses the MIMIC-III database, encompassing anonymized health records for adult patients aged 16 years and older admitted to critical care units between 2001 and 2012 \cite{bib15}. By leveraging this extensive dataset, our investigation aims to identify influential factors, enhance predictive models, and optimize clinical decision-making processes. These endeavors are aimed at mitigating in-hospital mortality rates linked to sepsis through the application of machine learning methodologies. The wide-ranging use of machine learning models in the healthcare field has shown great promise and significantly aided clinicians in making more informed decisions \cite{bib11,bib12,bib13}. Random Forest, a versatile ensemble learning technique, has proven particularly advantageous in the field of healthcare for predicting in-hospital mortality. One key benefit of Random Forest is its ability to handle high-dimensional data, which is typical in medical datasets \cite{bib16}. By averaging the results of multiple decision trees, Random Forest reduces the risk of overfitting and enhances predictive accuracy \cite{bib17}. Additionally, it provides feature importance scores, which are crucial for identifying significant clinical indicators of mortality. Recent studies have demonstrated the efficacy of Random Forest in this domain, showcasing its superior performance compared to other models. \newline

This research stands out by developing a highly accurate and interpretable machine learning model tailored for clinical environments. Utilizing advanced data processing techniques, including resampling and Random Forest feature importance, the model employs a compact Random Forest algorithm. This approach aims to assist healthcare professionals in optimizing resources and facilitating the timely assessment of sepsis patients. Additionally, several other machine learning models were used to ensure comprehensive analysis and validation of the results. \newline

\section{Methodology}

\subsection*{Data source and inclusion criteria}
We used the publicly available MIMIC-III database, which includes de-identified clinical data from patients admitted to the Beth Israel Deaconess Medical Center in Boston, Massachusetts. The MIMIC-III dataset contains records from 38,597 adult patients and 49,785 hospital admissions \cite{bib15}. It includes various tables with information such as admission details, patient demographics, caregiver data, lab results, charted observations, discharge summaries, and diagnosis codes. \newline

We selected our target patients from the MIMIC-III dataset based on the following criteria: (1) Patients aged 18 or older. (2) Patients diagnosed with sepsis. (3) Each patient is treated as a sample in the dataset. Ultimately, the total number of patients selected was 4,683, with admission counts totaling 17,429. \newline

\begin{figure}[!ht]
\centering
\includegraphics[width=0.6\textwidth]{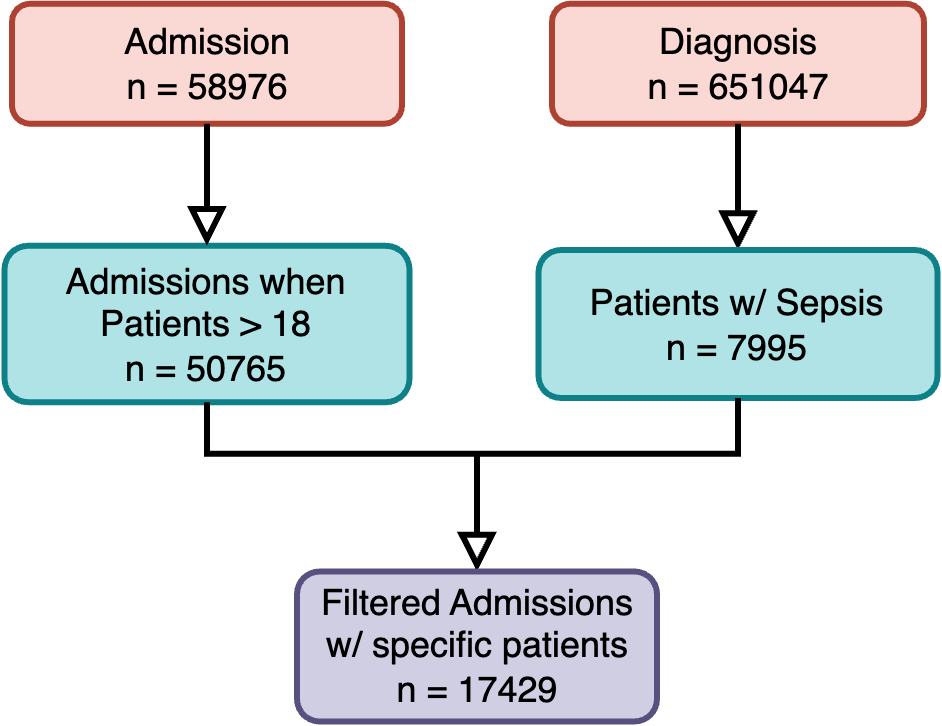} 
\caption{{\bf Patient Selection Process} 
Graphical representation of patient inclusion criteria.}
\label{fig1}
\end{figure}

For criterion (1), we calculated the patient's current age by subtracting the date of birth from the date of admission. If the date of admission was not recorded, we used the date of death instead. For criterion (2), we used several ICD-9 codes to identify symptoms of sepsis (995.91), severe sepsis (995.92), and septic shock (785.52). Patients were included in the study if any of these ICD-9 codes appeared in their most recent admission, following the standard methodology established in the existing literature \cite{bib18}. We then filtered out all patients who had these symptoms at least once. For criterion (3), we aggregated measurements and indicators in admission histories using different aggregation functions, including minimum, maximum, median, and average, to create a final table with one row per patient. A graphical expression for criteria selection is provided in Fig~\ref{fig1}. \newline

\begin{figure}[!ht]
\centering
\includegraphics[width=1\linewidth]{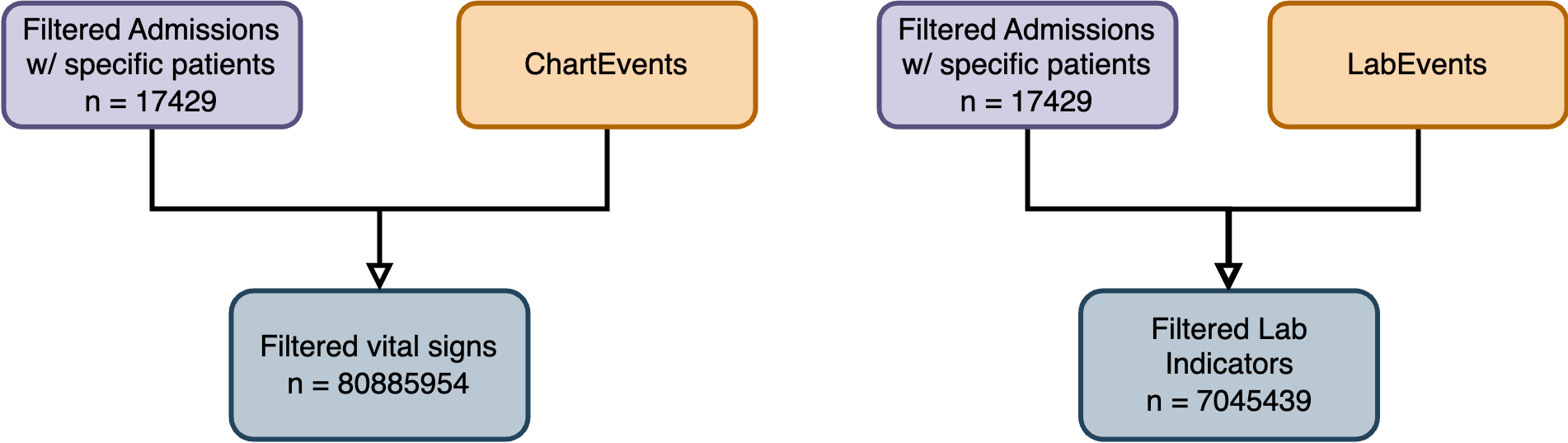} 
\newline
\caption{{\bf Data Extraction Process}
Graphical representation of health data and lab indicators Extraction}
\label{fig2}
\end{figure}

Data extraction followed the filtering of patients based on the above criteria. We extracted all health data and lab indicators from the ChartEvents and LabEvents tables. Initially, we loaded the dataset and excluded events unrelated to the target patients to reduce the data size for efficiency. A graphical expression for data extraction is provided in Fig~\ref{fig2}. \newline

\begin{figure}[!ht]
\centering
\includegraphics[width=0.8\linewidth]{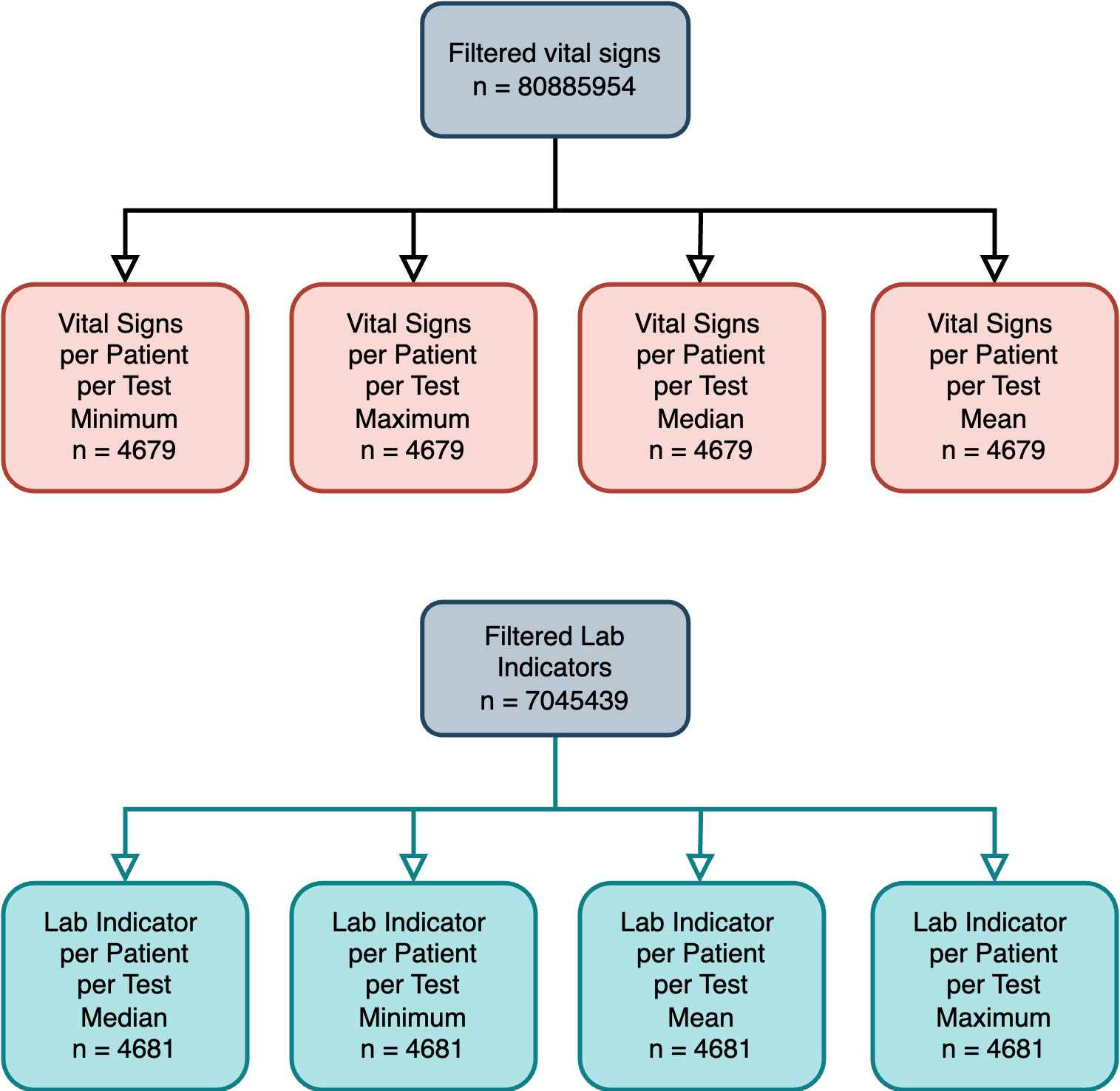} 
\newline
\caption{{\bf Aggregation and Pivoting Process}
Graphical representation of data aggregation and pivoting for patient features}
\label{fig3}
\end{figure}

The filtered ChartEvents and LabEvents tables were aggregated by each patient and each test (feature) using aggregation functions, including minimum, maximum, average, and median. Then, the table was pivoted so that each row represented one patient, with columns corresponding to the test (feature) IDs. A graphical expression for the aggregation process is provided in Fig~\ref{fig3}. \newline

\subsection{Data preprocessing}
Data preprocessing consisted of data cleaning, data imputation, and data splitting. Data cleaning included addressing missing values. Features with 30\% or more missing values were dropped from the dataset. The remaining missing values were imputed with the features' means after data splitting. The data was then split into two sets: training and testing, in a ratio of 75:25, respectively. Categorical values such as "ethnicity" and "gender" were encoded using the one-hot encoding technique. Scaling was applied to ensure all features were on a similar scale so the model could process them efficiently. Synthetic samples were created in the scaled feature space using the Synthetic Minority Over-sampling Technique (SMOTE) to address class imbalance, maintaining the integrity of the data distribution. Scaling and oversampling techniques were only applied to the training data to avoid data leakage \cite{bib19}. \newline

\subsection{Feature selection and feature importance}
% Fig~\ref{fig1}Table~\ref{table1}. \newline

\begin{figure}[!ht]
\centering
\includegraphics[width=1\linewidth]{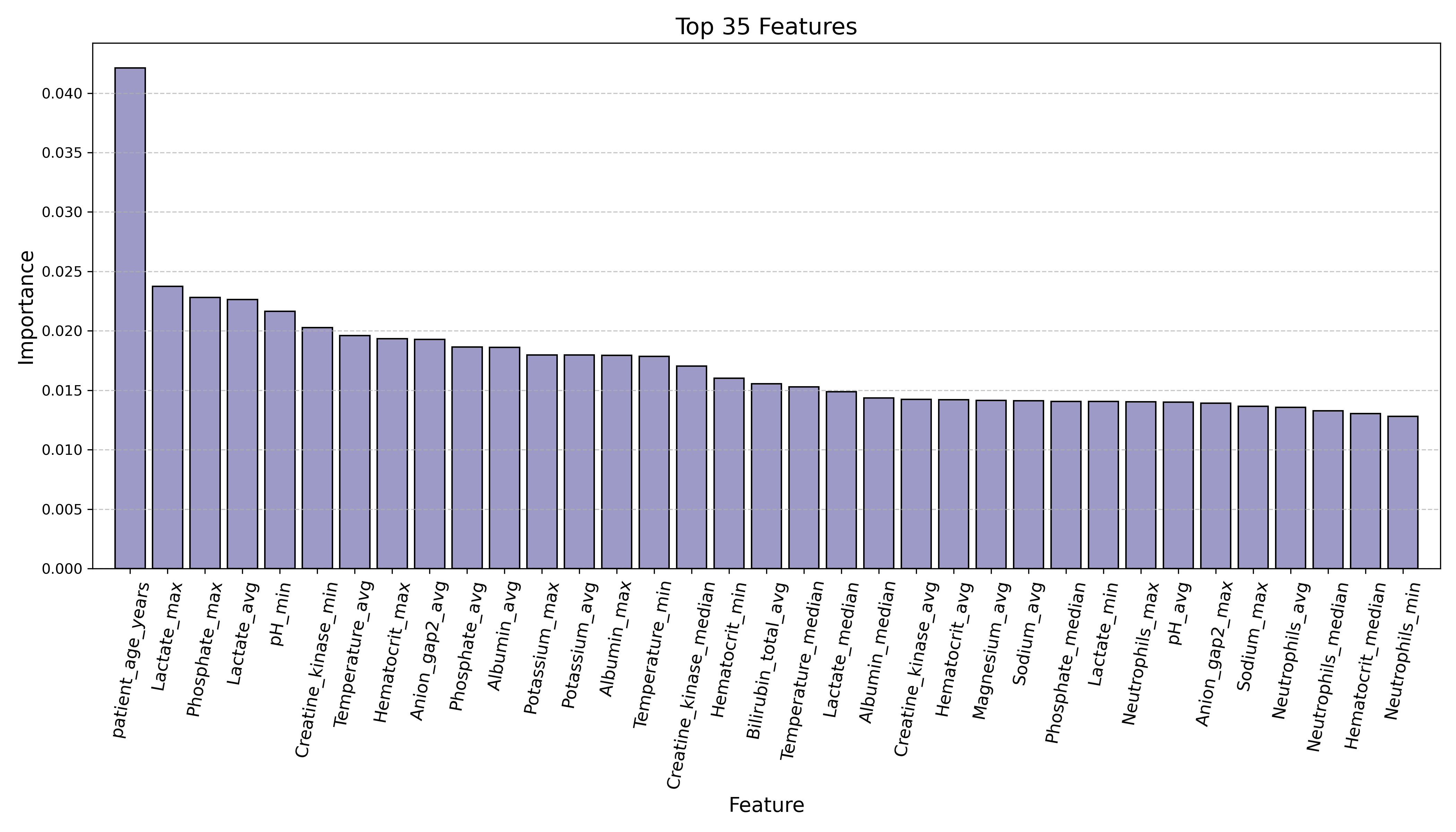} 
\caption{{\bf Feature Importance}
Graphical Representation of the Top 35 Most Important Features Identified by a Random Forest}
\label{fig4}
\end{figure}

The feature selection was a three-step process. First, a thorough literature review was conducted to select the initial 47 predictors of sepsis mortality. These features were included as the baseline indicators for future analysis. Second, based on our communication with medical experts, we included additional features crucial for studying in-hospital mortality due to sepsis. These features fell into the categories of vital signs, patient characteristics, and laboratory indicators. Finally, we ranked the refined predictors by their importance. Since one of the proposed and most promising models was Random Forest, the model's built-in feature importance attribute was used to measure each variable's impact on the prediction \cite{bib20}. The built-in feature importance attribute was chosen over other measures due to its simplicity and global interpretability. This attribute provides a high-level understanding of which features generally influence the model's predictions across the entire dataset. Another advantage is its fast computation, which is efficient when working with large datasets \cite{bib21}. The full list of these 35 features is provided in Table~\ref{tab:1}. The feature importance of the top 35 features is illustrated in Fig~\ref{fig4}. \newline

\begin{table}[!ht]
\centering
\caption{List of the Selected Features}
\label{tab:1}
\small
\renewcommand{\arraystretch}{1.5} % Adjust row height
\begin{tabular}{|>{\centering\arraybackslash}p{0.35\textwidth}|>{\centering\arraybackslash}p{0.33\textwidth}|>{\centering\arraybackslash}p{0.28\textwidth}|}
\hline
Age                               & Lactate max                     & Lactate avg                     \\ \hline
Phosphate max                     & Creatine Kinase min             & Albumin avg                     \\ \hline
pH min                            & Temperature avg                 & Creatine Kinase median          \\ \hline
Phosphate avg                     & Hematocrit max                  & Albumin max                     \\ \hline
Lactate median                    & Temperature min                 & Potassium max                   \\ \hline
Bilirubin, Total avg              & Hematocrit min                  & Creatine Kinase avg             \\ \hline
Albumin median                    & Sodium avg                      & Anion Gap max                   \\ \hline
Neutrophils median                & Magnesium avg                   & Hematocrit avg                  \\ \hline
Neutrophils max                   & Lactate Dehydrogenase max       & Temperature median              \\ \hline
Potassium, Whole Blood avg        & Lactate min                     & Hematocrit median               \\ \hline
Temperature avg                   & Creatine Kinase median          & Phosphate avg                   \\ \hline
Hematocrit max                    & Albumin max                     &                                \\ \hline
\end{tabular}
\end{table}

\subsection{Model development and optimization}

After data cleaning and feature selection, the modeling dataset comprised 8,690 observations. To comprehensively evaluate the performance of different machine learning classification models, a combination of techniques, including stratified train-test split, five-fold cross-validation, and SMOTE, was used. Since the class distribution was imbalanced, SMOTE helped mitigate this issue by generating synthetic examples for the minority class. The resulting dataset was then used to train five models: Logistic Regression, Gradient Boosting, Support Vector Machine (SVM), K-Nearest Neighbors (KNN), and Random Forest. Logistic Regression is a linear model used for binary classification that estimates probabilities using a logistic function, making it effective for predicting binary outcomes \cite{bib22}. Gradient Boosting is an ensemble technique that builds multiple decision trees sequentially, where each tree attempts to correct the errors of the previous one, resulting in a highly accurate model \cite{bib23}. SVM finds the optimal hyperplane that maximizes the margin between different classes, making it powerful for classification tasks with clear margins of separation \cite{bib28}. KNN classifies a sample based on the majority class of its nearest neighbors, using distance metrics to find the closest points, which makes it simple and intuitive for classification \cite{bib29}. Random Forest is an ensemble method that constructs multiple decision trees during training and outputs the mode of their predictions, combining the strengths of various trees to enhance performance and reduce overfitting \cite{bib30,bib31}. \newline

\begin{figure}[!ht]
\centering
\includegraphics[width=0.6\linewidth]{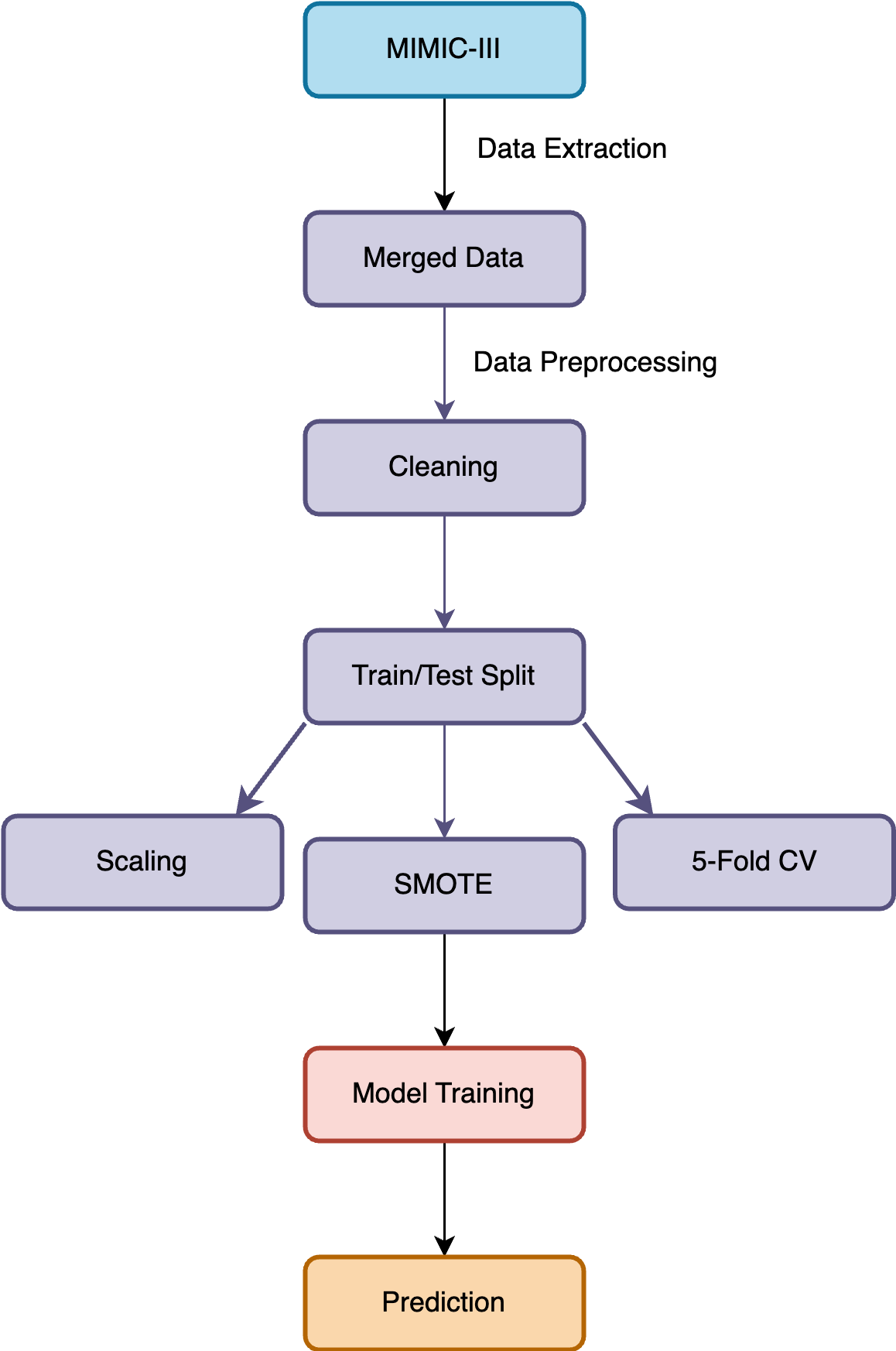} 
\caption{{\bf Data Processing and Model Training Workflow}
Graphical representation of the steps from data extraction to the model prediction}
\label{fig5}
\end{figure}

To select the most suitable model, we conducted a thorough evaluation of two key metrics: accuracy scores and AUROC scores. To assess the models’ stability, quantify uncertainty, and avoid overfitting, we used 95\% confidence intervals. After careful consideration, we decided to use the Random Forest model, which includes an in-built feature importance attribute. This decision was based on the model's superior AUROC scores, which indicate its ability to make accurate predictions and better discriminate between positive and negative outcomes, a critical aspect in the medical field. The workflow of the entire process is illustrated in Fig~\ref{fig5}.

\subsection{Statistical analysis between cohors}

A two-sided t-test statistical analysis was conducted to compare the variables' measurements in the train and test cohorts. The goal was to compare the means of each feature in the two cohorts to determine if there were statistically significant differences. By examining the p-values, which indicate the probability that the two sets have the same mean, an informed decision can be made about the validity of the model's assumptions. A standard threshold for considering differences as statistically significant is 0.05. Therefore, if many features have p-values below this threshold, it suggests significant differences between the train and test distributions for those features.

\section{Results}
\subsection*{Cohort characteristics model completion}
A two-sided t-test was conducted between corresponding features in the train and test cohorts. Each p-value tests the null hypothesis that each feature's train and test sets have identical average values. The obtained results indicate that there is not enough evidence to reject the null hypothesis. This means there is no significant difference between the mean values of the features in the training and test sets. This is a desirable outcome because it implies that each feature's training and test data are similarly distributed. The detailed cohort values and p-values reflecting differences between the training and testing sets are presented in Table~\ref{tab:2} \newline

\begin{table}[!ht]
\centering
\caption{Detailed overview of cohort characteristics for train and test cohort. Values are presented as means with the standard deviations in parentheses.}
\small % Make all fonts one size smaller
\begin{tabular}{|>{\centering\arraybackslash}p{0.27\textwidth}|>{\centering\arraybackslash}p{0.23\textwidth}|>{\centering\arraybackslash}p{0.23\textwidth}|>{\centering\arraybackslash}p{0.1\textwidth}|>{\centering\arraybackslash}p{0.1\textwidth}|}
\hline
\textbf{Characteristics} & \textbf{Train Cohort} & \textbf{Test Cohort} & \textbf{T-Stat} & \textbf{P-Value} \\
\hline
patient\_age & 64.152(15.196) & 63.918(15.390) & 0.594 & 0.552 \\
Lactate\_max & 5.054(3.748) & 5.010(3.767) & 0.453 & 0.650 \\
Phosphate\_max & 6.454(2.761) & 6.416(2.676) & 0.556 & 0.578 \\
Lactate\_avg & 2.455(1.729) & 2.425(1.754) & 0.680 & 0.497 \\
pH\_min & 7.205(0.184) & 7.213(0.131) & -2.024 & 0.043 \\
Creatine\_kinase\_min & 129.529(1583.876) & 91.565(372.818) & 1.729 & 0.084 \\
Temperature\_avg & 37.042(0.654) & 37.038(0.639) & 0.247 & 0.805 \\
Hematocrit\_max & 40.020(5.679) & 40.045(5.671) & -0.173 & 0.863 \\
Anion\_gap2\_avg & 14.205(2.979) & 14.123(3.031) & 1.055 & 0.292 \\
Phosphate\_avg & 3.642(0.948) & 3.610(0.962) & 1.276 & 0.202 \\
Albumin\_avg & 2.969(0.539) & 2.952(0.535) & 1.197 & 0.231 \\
Potassium\_max & 5.981(1.404) & 6.026(1.440) & -1.227 & 0.220 \\
Potassium\_avg & 4.143(0.358) & 4.138(0.350) & 0.583 & 0.560 \\
Albumin\_max & 3.630(0.717) & 3.630(0.731) & -0.019 & 0.985 \\
Temperature\_min & 36.020(1.152) & 36.011(1.224) & 0.301 & 0.763 \\
Creatine\_kinase\_median & 358.224(2934.465) & 261.064(979.051) & 2.234 & 0.026 \\
Hematocrit\_min & 23.178(4.886) & 23.017(4.770) & 1.314 & 0.189 \\
Bilirubin\_total\_avg & 1.966(3.978) & 1.794(3.596) & 1.811 & 0.070 \\
Temperature\_median & 37.034(0.672) & 37.028(0.651) & 0.342 & 0.732 \\
Lactate\_median & 2.195(1.732) & 2.174(1.774) & 0.467 & 0.640 \\
Albumin\_median & 2.944(0.576) & 2.923(0.567) & 1.406 & 0.160 \\
Creatine\_kinase\_avg & 450.048(3063.969) & 331.290(1133.093) & 2.543 & 0.011 \\
Hematocrit\_avg & 30.597(3.477) & 30.469(3.422) & 1.460 & 0.144 \\
Magnesium\_avg & 2.021(0.215) & 2.013(0.207) & 1.583 & 0.113 \\
Sodium\_avg & 138.650(3.465) & 138.588(3.271) & 0.729 & 0.466 \\
Phosphate\_median & 3.532(0.943) & 3.502(0.963) & 1.206 & 0.228 \\
Lactate\_min & 1.156(0.997) & 1.145(1.060) & 0.391 & 0.696 \\
Neutrophils\_max & 88.851(9.514) & 89.436(8.341) & -2.636 & 0.008 \\
pH\_avg & 7.364(0.064) & 7.366(0.062) & -0.933 & 0.351 \\
Anion\_gap2\_max & 22.469(6.259) & 22.444(6.382) & 0.153 & 0.879 \\
Sodium\_max & 146.455(5.418) & 146.414(5.034) & 0.306 & 0.760 \\
Neutrophils\_avg & 75.895(12.170) & 76.174(11.405) & -0.937 & 0.349 \\
Neutrophils\_median & 76.672(13.004) & 76.833(12.391) & -0.498 & 0.619 \\
Hematocrit\_median & 30.282(3.564) & 30.139(3.506) & 1.583 & 0.114 \\
Neutrophils\_min & 59.057(20.488) & 58.880(20.251) & 0.340 & 0.734 \\
\hline
\end{tabular}
\label{tab:2}
\end{table}

\subsection{Evaluation metrics proposed and baseline models’ performance}

Table~\ref{tab:3} presents the results of different models. It includes a detailed evaluation of the models' performances, such as AUROC score, precision, sensitivity, accuracy, and F1 score. Among these models, the Random Forest model achieved the best results, with an AUROC score of 0.97.

\begin{table}[!ht]
\centering
\caption{Summary of the evaluation metrics for the prediction models on the test set}
\small
\renewcommand{\arraystretch}{1.2} % Adjust row height for better readability
\setlength{\tabcolsep}{5pt} % Adjust column spacing
\begin{tabular}{|c|l|c|c|c|c|}
\hline
\textbf{Model} & \textbf{AUROC (95\% CI)}        & \textbf{Accuracy} & \textbf{Precision} & \textbf{Recall} & \textbf{F-Score} \\
\hline
RF  & 0.9715 [0.9656 - 0.9769] & 0.9003         & 0.93             & 0.91          & 0.92       \\
GB  & 0.8652 [0.8484 - 0.8808] & 0.7901         & 0.87             & 0.79          & 0.83       \\
LR  & 0.7701 [0.7491 - 0.7899] & 0.7007         & 0.81             & 0.69          & 0.75       \\
KNN & 0.8840 [0.8687 - 0.8990] & 0.7793         & 0.90             & 0.74          & 0.81       \\
SVM & 0.8628 [0.8453 - 0.8790] & 0.7827         & 0.87             & 0.77          & 0.82       \\
\hline
\end{tabular}
\label{tab:3}
\end{table}

\begin{figure}[!ht]
\includegraphics[width=1\linewidth]{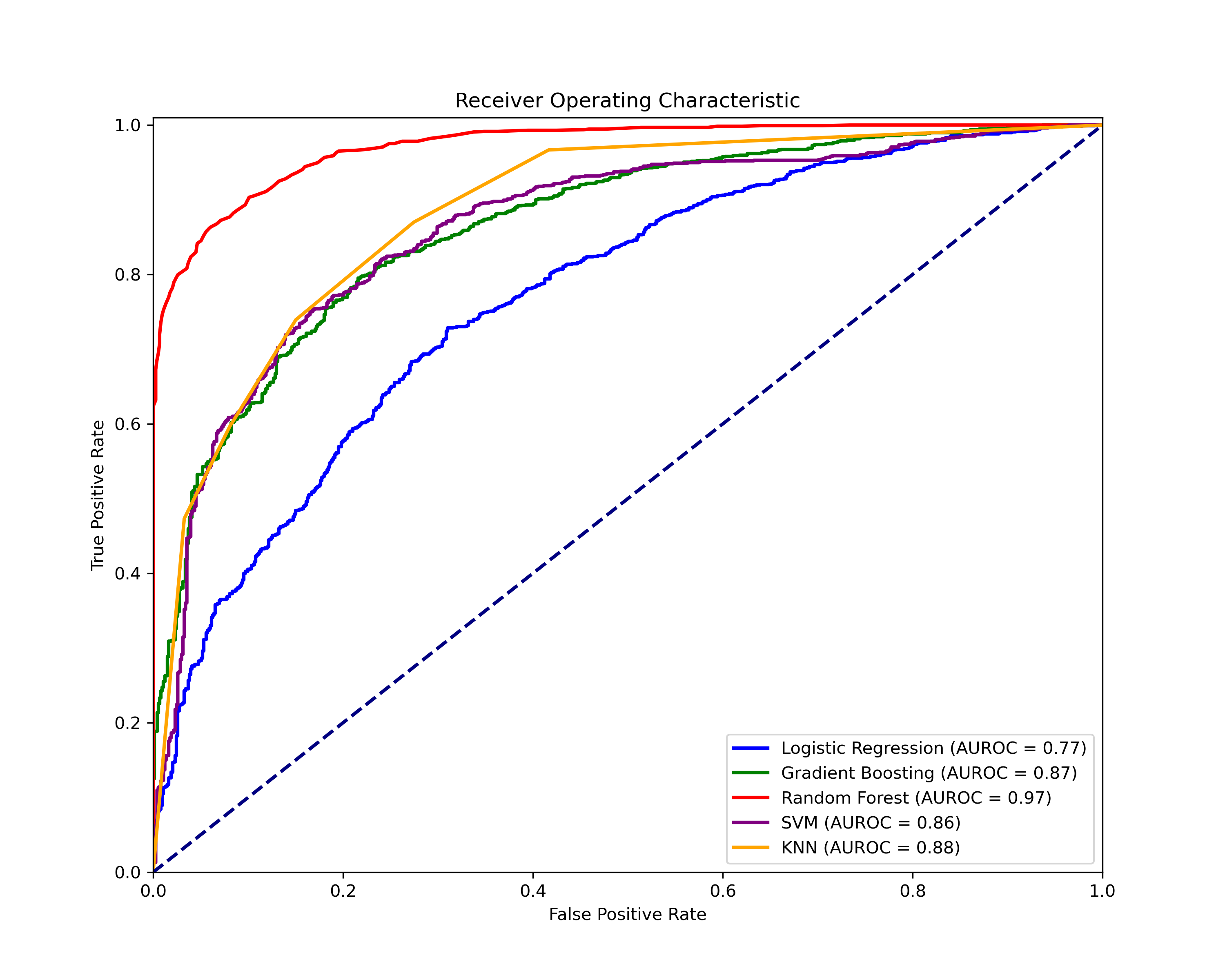} 
\caption{{\bf Model Comparison}
The AUROC curves and scores for five models: Logistic Regression, Gradiant Boosting, Random Forest, SVM, and KNN.}
\label{fig6}
\end{figure}

The Receiver Operating Characteristic (ROC) curves presented in Fig~\ref{fig6} display the performance of the five machine learning models in predicting the outcome of in-hospital sepsis mortality. The models evaluated include Logistic Regression, Gradient Boosting, Random Forest, SVM, and KNN. The Random Forest model demonstrates superior performance with an AUROC of 0.97, indicating excellent discriminative ability. The KNN model follows with an AUROC of 0.88, closely matched by the Gradient Boosting and SVM models, both with an AUROC of 0.86. The Logistic Regression model performs comparatively less, with an AUROC of 0.77. These findings suggest that the Random Forest model is the most effective for the predictive task, significantly outperforming the other models in balancing the true positive rate and false positive rate. 

\begin{figure}[!ht]
\includegraphics[width=1\linewidth]{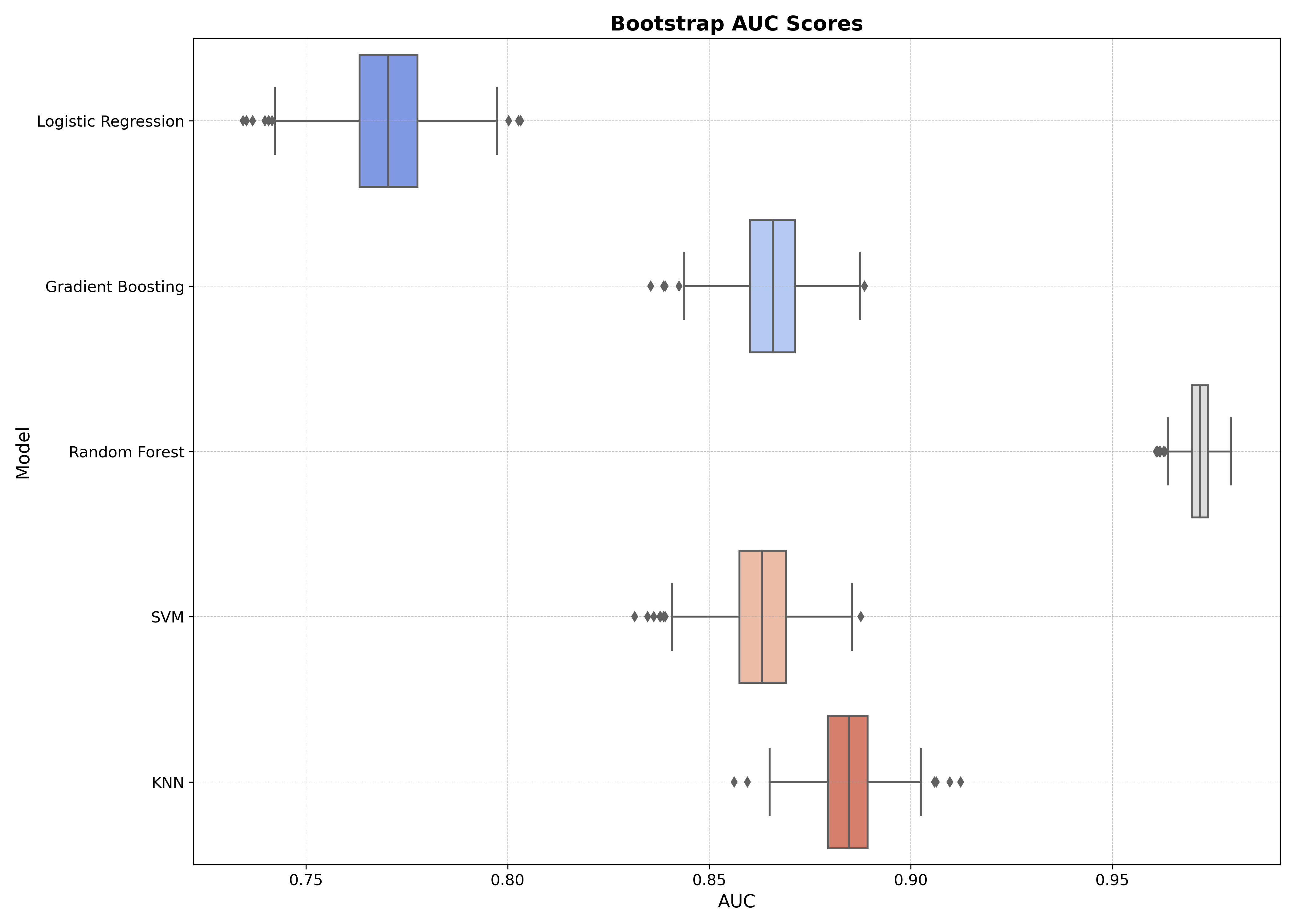} 
\caption{{\bf Model Comparison}
The bootstrap AUC scores for five models: Logistic Regression, Gradient Boosting, Random Forest, SVM, and KNN. The box plot shows the distribution of AUC scores obtained from bootstrap sampling.}
\label{fig7}
\end{figure}

The boxplot visualization in Fig~\ref{fig7} displays the distribution of AUC scores for five machine learning models using bootstrap resampling. The Random Forest model exhibits the highest median AUC and a relatively narrow interquartile range, signifying both high performance and consistency. The KNN and SVM models have similar median AUC values, both slightly lower than the Random Forest, but display wider interquartile ranges, indicating more variability in their performance. Gradient Boosting shows a median AUC comparable to KNN and SVM but with tighter variability. Logistic Regression, in contrast, demonstrates the lowest median AUC and the widest distribution, suggesting it is less effective and more inconsistent than the other models.

\subsection{Shapley Value analysis}
\begin{figure}[!ht]
\includegraphics[width=0.85\linewidth]{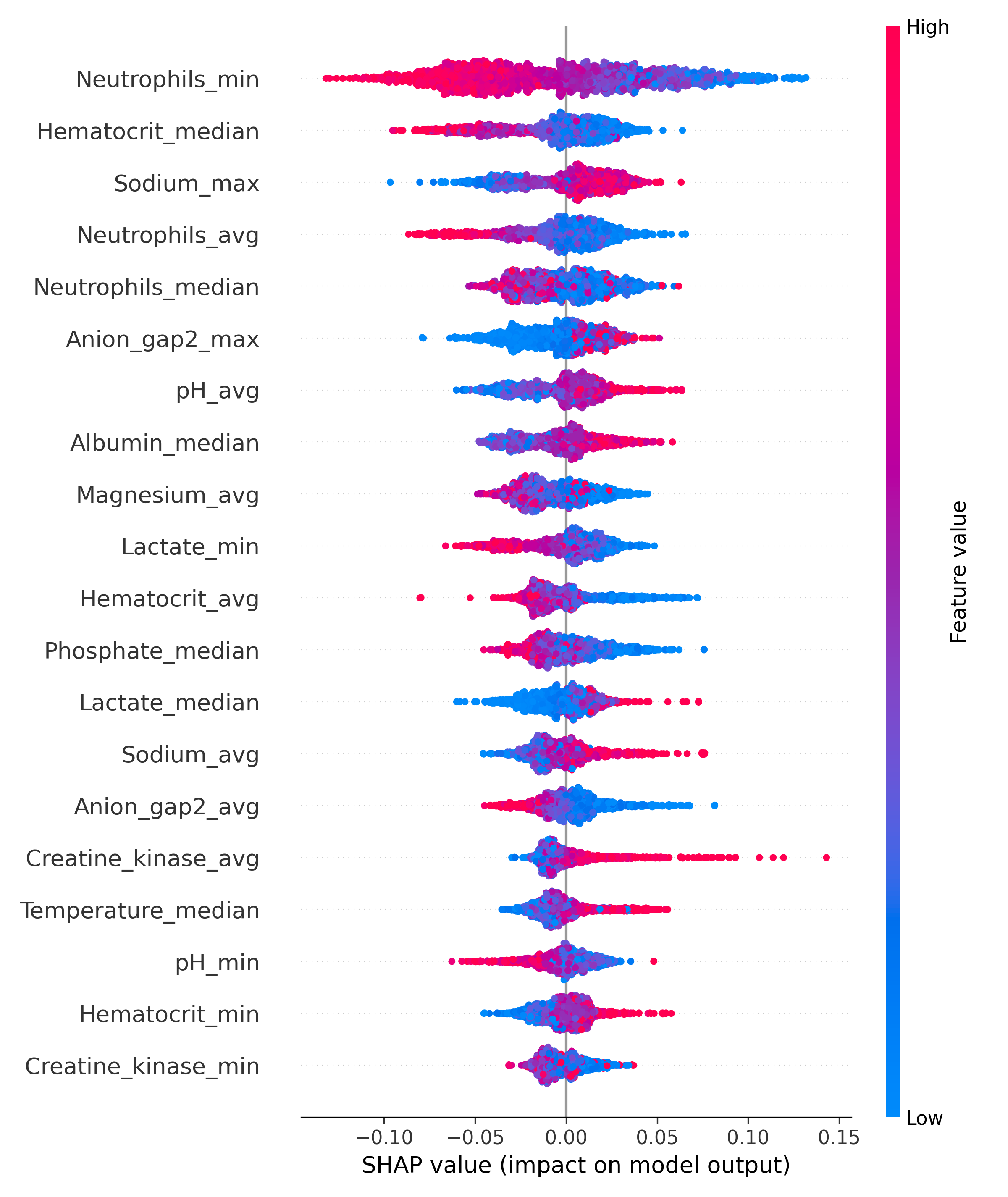} 
\centering
\caption{{\bf SHAP Analysis}
Summary plot for top feature impacts in sepsis prediction based on the Random Forest model.}
\label{fig8}
\end{figure}

\begin{figure}[!ht]
\includegraphics[width=0.85\linewidth]{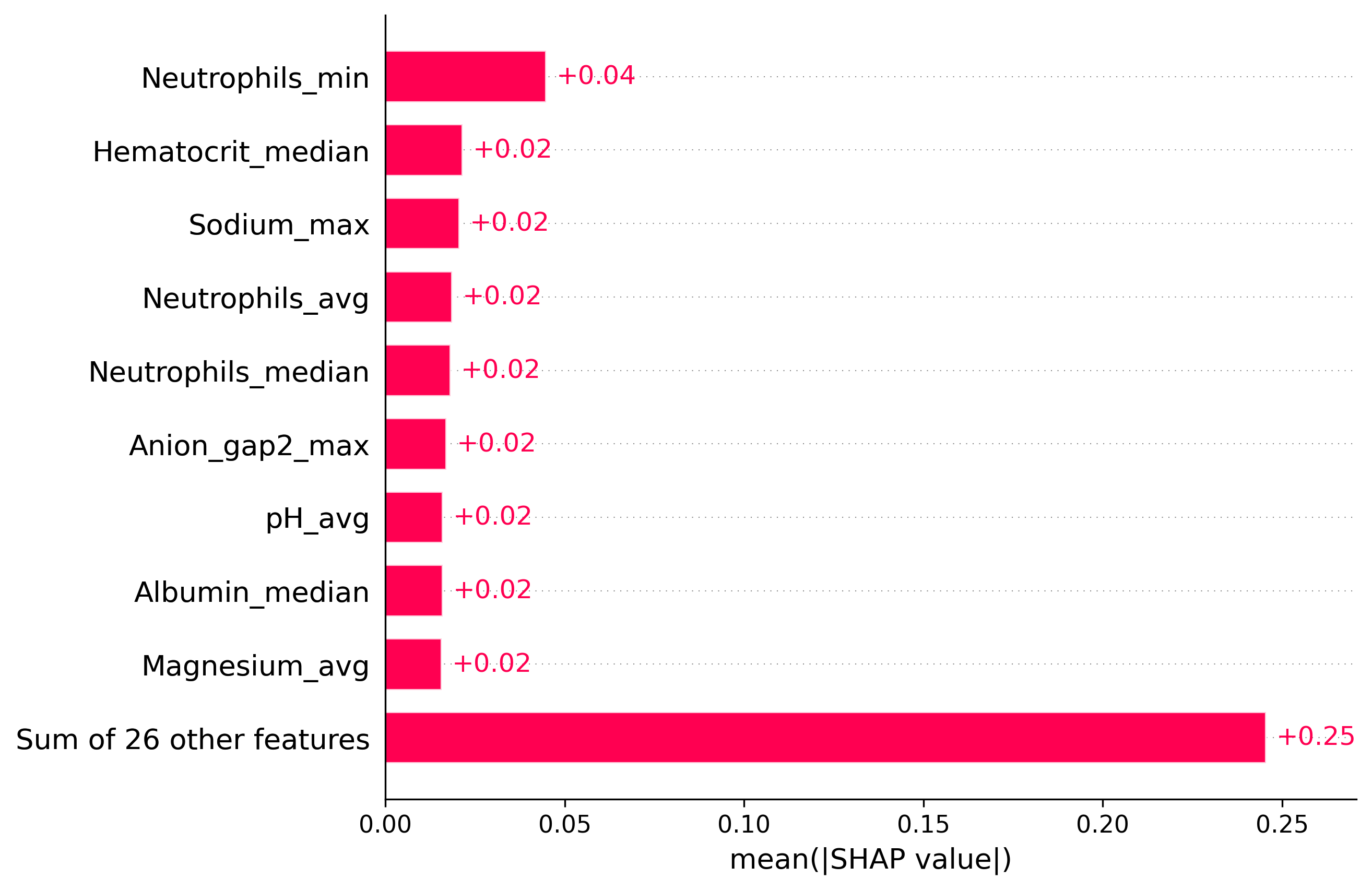} 
\caption{{\bf SHAP Analysis}
Bar plot of mean absolute SHAP values, demonstrating the
impact of each feature on the model’s prediction}
\label{fig9}
\end{figure}

As shown in Fig~\ref{fig8}, SHAP analysis of the Random Forest classifier identified "Neutrophils\_min“, "Hematocrit\_median,” “Sodium\_max” and “Neutrophils\_avg” as the most influential features in predicting in-hospital sepsis mortality. These results align with recent research literature, underscoring the significance of hematocrit and lactate levels in predicting sepsis mortality \cite{bib32,bib33}. Although the SHAP summary plot typically displays the top 20 features for clarity, all 35 features were analyzed. The order of the predictors in the SHAP summary plot differs from the Random Forest’s in-built “feature\_importance” attribute. Such a discrepancy is expected because SHAP values consider the feature’s contribution in the context of specific predictions rather than providing a global measure of importance. SHAP values provide a more nuanced view of feature interactions. The analysis offers valuable insights into the proposed model’s decision-making process and highlights the significance of individual feature interactions. Additionally, the absolute mean SHAP values plot (Fig~\ref{fig9}) demonstrates that the same predictors had the highest average impact. This plot further supports the significance of these features in the model's predictions.

\section{Discussion}
\subsection{Summary of existing model compilation}

Over the past few years, machine learning and deep learning models have emerged as promising predictive solutions \cite{bib34,bib35,bib37}. These advanced algorithms have been applied across various domains, including healthcare, to predict patient outcomes and improve clinical decision-making. For example, Yong and Zhenzhou proposed a deep learning mortality risk assessment model for sepsis patients \cite{bib18}. Similarly, Bao et al. showcased the significance of the Light GBM algorithm in predicting sepsis patient mortality, comparing the effectiveness of several machine learning models \cite{bib38}. However, despite utilizing advanced analytical techniques, these studies have yet to achieve high predictive results that are easy to interpret. This study also serves as our primary point of comparison. \newline

To further enhance our understanding of the subject and refine feature selection for improved accuracy, we conducted an exhaustive review of pertinent literature on sepsis mortality rates. Additionally, we deepened our exploration of medical insights into sepsis and examined diverse feature selection methodologies aimed at optimizing parameter selection for predicting sepsis mortality. For instance, Ye et al. highlight elevated organ dysfunction scores, reduced Body Mass Index, body temperature variations, heightened heart rate, and decreased urine output as pivotal indicators of sepsis mortality \cite{bib36}. \newline

Table~\ref{tab:4} compares this research with the best existing study that predicted sepsis mortality. In their study, Yong and Zhenzhou also used the MIMIC-III database to extract features and utilized similar criteria in selecting patient data \cite{bib18}. They applied the same exclusion criteria, dropping observations with more than 30\% missing values. Their best model was a deep learning model called DGFSD, but our proposed model, Random Forest, showed better accuracy results. Most importantly, our study reported an AUROC score, which is more crucial in the clinical setting because it measures a model's ability to discriminate between classes, distinguishing between positive cases (true positives) and negative cases (true negatives). It effectively quantifies how well a model can differentiate between true positive rates and false positive rates. Moreover, our model is less complex and more straightforward to interpret. Therefore, our proposed model outperforms the existing literature in its predictive capabilities for in-hospital sepsis mortality.

\begin{table}[!ht]
\centering
\caption{Performance Comparison}
\begin{tabular}{|c|c|c|}
\hline
\textbf{} & \textbf{Yong, L., Zhenzhou, L. \cite{bib18}} & \textbf{This Study} \\ \hline
\textbf{Model} & Deep Learning & Random Forest \\ \hline
\textbf{Accuracy} & 0.82 & 0.90 \\ \hline
\textbf{AUROC} & Not reported & 0.97 \\ \hline
\end{tabular}
\label{tab:4}
\end{table}

\subsection{Study limitations and future research}

This study has some limitations despite significant progress. One limitation is that MIMIC-III was used for data extraction, while a newer version of the database, MIMIC-IV, a contemporary electronic health record dataset covering a decade of admissions between 2008 and 2019, is already available \cite{bib39}. In the US, nearly 96\% of hospitals had a digital electronic health record system (EHR) in 2015, the records of which are included in the MIMIC-IV database, with more accurate and reliable data. Therefore, it would be a good practice for future studies to utilize the latest data available. Another limitation is that machine learning techniques are relatively new and complex data analysis methods; thus, their results can be easier to interpret with enough theoretical background. Additionally, this study has not yet considered other more elaborate techniques, such as deep learning. Along with machine learning, deep learning methods have been proven efficient by other studies in the medical field.

\section{Conclusion}
The study has achieved substantial advancements in predicting sepsis mortality by employing sophisticated machine learning techniques. These methods, along with the preprocessing strategies chosen, effectively mitigate data imbalance issues inherent in the MIMIC-III database. They also utilize a carefully selected set of features to produce highly accurate predictions, as demonstrated by the achieved AUROC score. The incorporation of the Random Forest model’s built-in feature importance attribute allows for detailed and easily understandable explanations of feature relevance, making the model interpretable for clinicians and a wide range of audiences. Consequently, our model is not only straightforward to interpret and highly accurate but also surpasses the predictive capabilities of existing models.

The findings of our research underscore the exceptional performance of the Random Forest model, which attained an AUROC score of 0.97, precision of 0.93, recall of 0.91, accuracy of 0.90, and an F1 score of 0.92. This research showcases the potential of machine learning to enhance decision-making in critical care by employing advanced techniques to predict and prevent sepsis-related fatalities. The study proposes that integrating these predictive models into clinical workflows could transform patient care, providing healthcare professionals with a crucial tool to combat sepsis and reduce in-hospital sepsis mortality.


\end{document}